\documentclass[10pt,twocolumn,letterpaper]{article}

\usepackage[pagenumbers]{cvpr} 

\usepackage{graphicx}
\usepackage{amsmath}
\usepackage{amssymb}
\usepackage{booktabs}
\usepackage{epstopdf}  
\usepackage{multirow}
\usepackage{array}  
\usepackage{color}
\usepackage{colortbl}
\usepackage{mathrsfs}  
\usepackage{amsfonts}  
\usepackage[switch]{lineno}
\usepackage{algorithm}
\newcommand{\bm}[1]{\mbox{\boldmath{$#1$}}}
\usepackage{algpseudocode}
\usepackage{bm}
\usepackage{rotating}

\usepackage[pagebackref,breaklinks,colorlinks]{hyperref}

\usepackage[capitalize]{cleveref}
\crefname{section}{Sec.}{Secs.}
\Crefname{section}{Section}{Sections}
\Crefname{table}{Table}{Tables}
\crefname{table}{Tab.}{Tabs.}

\def\cvprPaperID{3827} 
\def\confName{CVPR}
\def\confYear{2022}

\begin{document}

\title{Robust Region Feature Synthesizer for Zero-Shot Object Detection}

\author{Peiliang Huang$^{1}$, Junwei Han$^{1}$, De Cheng$^{2}$, Dingwen Zhang$^{1}$\\
$^{1}$Brain and Artificial Intelligence Lab, School of Automation, Northwestern Polytechnical University\\
$^{2}$State Key Laboratory of Integrated Services Networks, School of Telecommunications Engineering, \\Xidian University\\
{\tt\small \{peilianghuang2017, junweihan2010, zhangdingwen2006yyy\}@gmail.com, dcheng@xidian.edu.cn}
}
\maketitle

\begin{abstract}
  Zero-shot object detection aims at incorporating class semantic vectors to realize the detection of (both seen and) unseen classes given an unconstrained test image. In this study, we reveal the core challenges in this research area: how to synthesize robust region features (for unseen objects) that are as intra-class diverse and inter-class separable as the real samples, so that strong unseen object detectors can be trained upon them. To address these challenges, we build a novel zero-shot object detection framework that contains an Intra-class Semantic Diverging component and an Inter-class Structure Preserving component. The former is used to realize the one-to-more mapping to obtain diverse visual features from each class semantic vector, preventing miss-classifying the real unseen objects as image backgrounds. While the latter is used to avoid the synthesized features too scattered to mix up the inter-class and foreground-background relationship. To demonstrate the effectiveness of the proposed approach, comprehensive experiments on PASCAL VOC, COCO, and DIOR datasets are conducted. Notably, our approach achieves the new state-of-the-art performance on PASCAL VOC and COCO and it is the first study to carry out zero-shot object detection in remote sensing imagery.

\end{abstract}

\vspace{-6mm}
\section{Introduction}
\label{sec:intro}
With the rapid development of the deep learning technologies, such as CNN \cite{ren2016faster,han2018advanced} and Transformer \cite{liu2021swin}, great progresses have been made in the research field of object detection. Although the detection performance achieved by existing methods looks promising and encouraging, it exists a hidden drawback for applying them in real-world scenarios---The mainstream detection approaches have the strict constraint on the category to detect. Once the model is trained, it can only recognize objects that appear in the training data, whereas other objects appearing in the test images but unseen during training would confuse the model dramatically, leading to avoidless faults in detection results. To address this problem, the task of zero-shot object detection (ZSD)\cite{bansal2018zero, rahman2018zero, zhu2020don, hayat2020synthesizing} was raised in recent years. The goal is to enable the detection models to predict unseen objects which are without any available samples during training.

Earlier efforts on zero-shot object detection (ZSD) \cite{bansal2018zero, rahman2018zero} focus on mapping function-based methods, which learn mapping functions from the visual space to the semantic space. With the learned mapping functions, unseen object categories can be predicted by mapping their visual features into the semantic space and then performing the nearest neighbor search in the semantic space. However, due to that the mapping functions are learned all upon the seen categories provided by the training data, the models would get significantly biased towards the seen categories when dealing with the visual features in testing \cite{hayat2020synthesizing}. Recently, generative model-based methods \cite{zhu2020don, hayat2020synthesizing} are presented as an alternative solution. Usually, these methods utilize generative models to synthesize visual features from the provided semantic embeddings \cite{mikolov2013distributed, akata2015evaluation} corresponding to each object category. The synthesized visual features can then be used for training a standard detector for unseen classes. Generative model-based methods show stronger performance compared with mapping function-based methods in solving the bias problem as, although the samples corresponding to the unseen objects are still absent, the detectors are trained with synthesized visual features for the unseen objects.

\begin{figure*}[t]
	\begin{center}
		\includegraphics[width=1\linewidth]{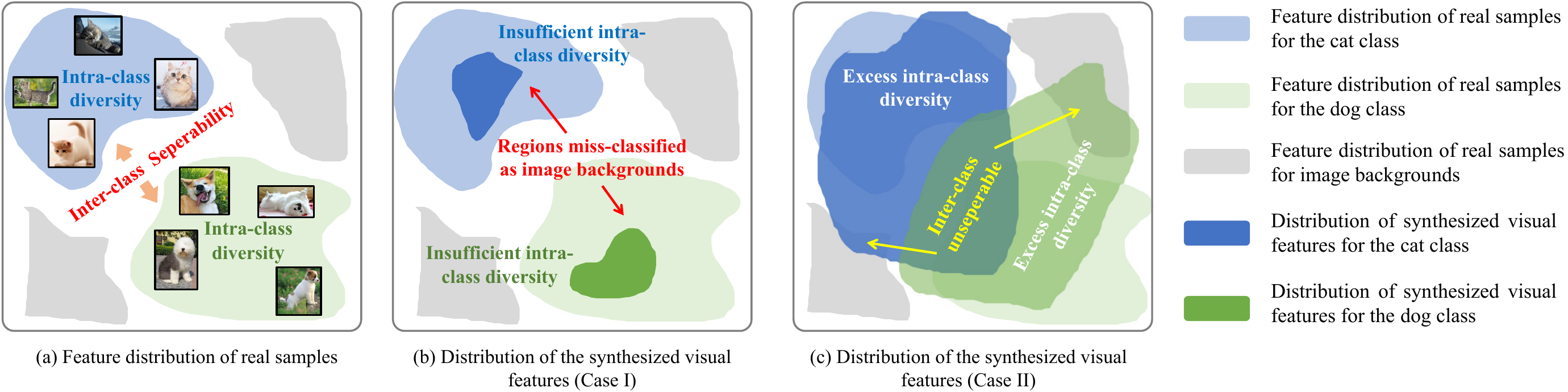}
	\end{center}
		\vspace{-3mm}
	\caption{Illustrations of the problem studied in this work. In real cases, the feature space built by the samples show high intra-class diversity but still with inter-class separability like in (a), whereas the spaces of the synthesized visual features learned by existing approaches either have insufficient intra-class diversity, as shown in (b), or have excess intra-class diversity to make inter-class inseparable, as shown in (c).}
	\label{fig_intro}
	\vspace{-3mm}
\end{figure*}

However, the current generative model-based methods mainly follow the ideas presented in zero-short classification frameworks, such as \cite{wu2020self,pambala2020generative}, where the synthesized visual features may perform well in the less complex classification scenarios but are not robust enough to obtain satisfying results in the complicated detection scenarios. To our best knowledge, there are two-fold challenges for synthesizing visual features for detection scenarios:
	\begin{itemize}
		\item \textbf{Intra-class diversity:} Objects in real-world detection scenarios present high variation in pose, shape, texture, etc., and one object instance may be covered by several bounding boxes with different sizes and locations. This leads to the high diversity in their feature representations.
		\item \textbf{Inter-class separability:} Though having such variations, each object category still has easy-to-recognized characteristics that are distinct from other object categories as well as the image backgrounds, making the feature representations from different classes (including the background class) highly separable.
	\end{itemize}
Although some existing approaches have recognized the importance of intra-class diversity \cite{zhao2020gtnet,hayat2020synthesizing}, without jointly considering the inter-class separability, these methods would either impose insufficient diversity to the synthesized visual features, leading to miss-classify the real unseen objects as image backgrounds (see Fig \ref{fig_intro} (b)), or go too far to make the visual features synthesized for different class semantics mixed together, thus making the learned detection models obtain inaccurate object categories for foreground regions or suffer from errors in dealing with the image backgrounds (see Fig \ref{fig_intro} (c)).



\begin{figure*}[t]
  \centering
  	\vspace{1.8mm}
   \includegraphics[width=1.0\linewidth]{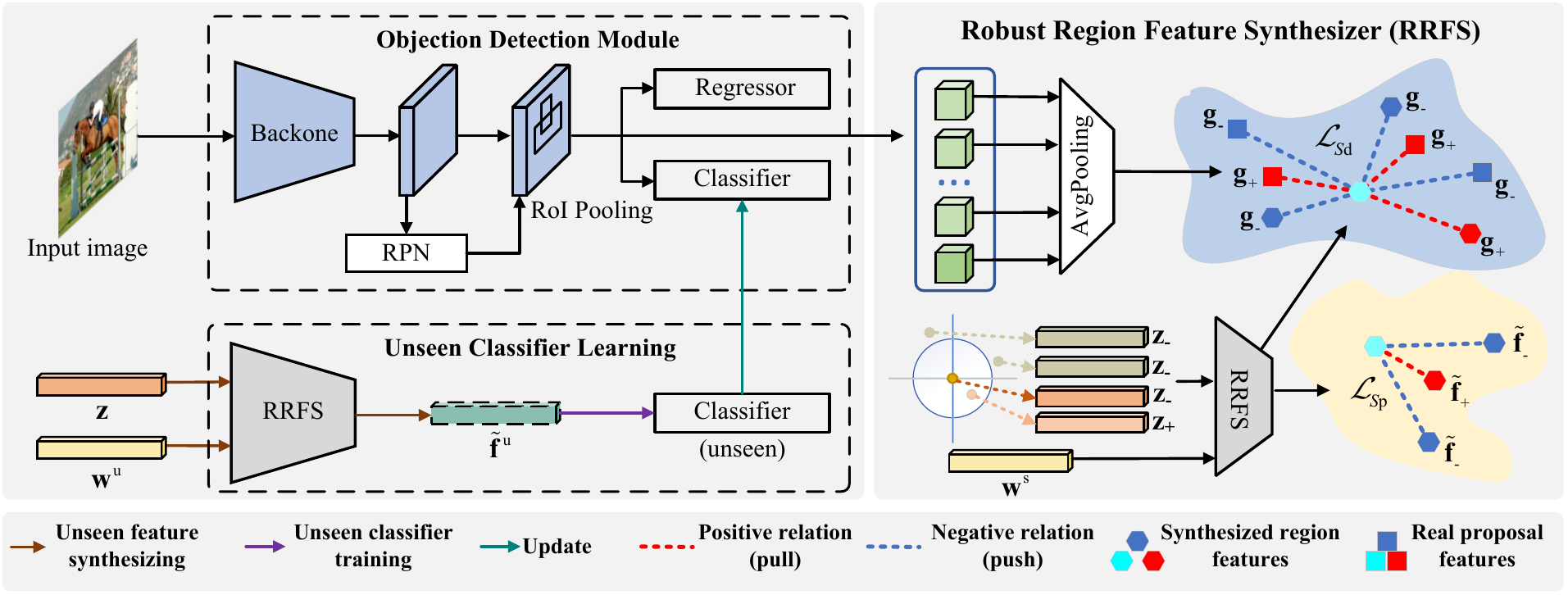}
   \caption{Illustration of the proposed overall framework. Our method contains an object detection  module and a unseen classifier learning module. The basic idea is to learn object detector based on the labeled seen category data firstly, and then use the synthesized unseen region features to train unseen classifiers. To keep the framework simple and easy to understand, we do not show the discriminator used in the learning process.}
   \label{framwork}
   	\vspace{-3mm}
\end{figure*}

To overcome the feature synthesizing problems toward real-world detection scenarios, we build a novel zero-shot object detection framework as shown in Fig~\ref{framwork}. Specifically, we design two components for learning robust region features. To enable the model to synthesize diverse visual features, we propose an Intra-class Semantic Diverging (IntraSD) component which can diverge the semantic vector of a single class into a set of visual features. To prevent the intra-class diversity of the synthesized features goes too far to mix up the inter-class relationship, we further propose an Inter-class Structure Preserving (InterSP) component that utilizes real visual samples from different object categories to constrain the separability of the synthesized visual features.

It is also worth mentioning that in the design of InterSP, we fully leverage the region features sampled from the real image scenes for detection instead of implementing it on the synthesized visual features. This enables our model to synthesize visual features as separable as in real cases and obtain much better performance when compared to the aforementioned counterpart (see experiments in Section ~\ref{ablation_study}).


To sum up, this paper mainly has the following three-fold contributions:
	\begin{itemize}
		\item We reveal the key challenges, i.e., the intra-class diversity and inter-class separability, for feature synthesizing in real-world object detection scenarios.
		\item With the goal to synthesize robust region features for ZSD, we build a novel framework that contains an Intra-class Semantic Diverging component and an Inter-class Structure Preserving component.
		\item Comprehensive experiments on three datasets, including PASCAL VOC, COCO, and DIOR, demonstrate the effectiveness of the proposed approach. Notably, this is also the first attempt for implementing zero-shot object detection in remote sensing imagery.
	\end{itemize}

\section{Related Work}
\label{sec:formatting}
\textbf{Zero-shot Learning (ZSL).} ZSL aims to use seen examples to train the network and reason about unseen classes by leveraging the semantic label embeddings (e.g. word-vector \cite{mikolov2013distributed} or semantic attributes \cite{akata2015evaluation}) as side information. Earlier ZSL research works focus on the embedding function-based methods, which embed the visual features into the semantic descriptor space, or vice versa \cite{akata2013label, fu2014transductive, bucher2016improving, kodirov2017semantic}. As a result, the visual features and the semantic features will lie in a same embedding space and the ZSL classification can be accomplished by searching the nearest semantic descriptor in the embedding space \cite{han2020learning}. Embedding function-based methods work well in conventional ZSL scenario \cite{akata2013label, bucher2016improving, xian2016latent, fu2015zero} but tend to be highly overfitting the seen classes in the more challenging GZSL scenario \cite{han2020learning, chen2021free, verma2018generalized, xian2018feature}. To tackle this overfitting problem, some researchers have introduced generative-based methods \cite{schonfeld2019generalized, kingma2013auto, han2020learning, chen2021free}, which learn to complement the training samples for unseen classes by
using a conditional generative model e.g. Variational Autoencoder (VAE) \cite{kingma2013auto} and Generative Adversarial Networks (GAN) \cite{xian2019f}. With the synthesized unseen classes examples, they can transform the zero-shot classification problem to a general fully-supervised problem and relieve the overfitting problem. In this paper, we also employ a generative model to synthesize unseen visual features for converting the ZSL into a fully supervised way. However, as our goal is to solve the more challenging ZSD problem, we need to handle heavier intra-class diversity and inter-class separability in model design.

\textbf{Zero-shot object detection.} ZSD receives great research interest in recent years \cite{rahman2018zero, bansal2018zero, demirel2018zero, li2019zero, zheng2020background, rahman2020improved, zhu2020don, zhao2020gtnet, hayat2020synthesizing}. Some researches focus on embedding function-based methods \cite{rahman2018zero, bansal2018zero, demirel2018zero, li2019zero, zheng2020background, rahman2020improved}. Unfortunately, these methods would suffer from the overfitting problem like in ZSL, where the unseen objects are significantly biased towards the seen classes or background \cite{hayat2020synthesizing}. Generative model-based methods \cite{zhu2020don, zhao2020gtnet, hayat2020synthesizing} show strong performance in solving the bias problem. Zhu \emph{et al.} \cite{zhu2020don} synthesized visual features for unseen objects from semantic information and augments existing training algorithms to incorporate unseen object detection. Zhao \emph{et al.} \cite{zhao2020gtnet} proposed a Generative Transfer Network for zero-shot object detection. Hayat \emph{et al.}\cite{hayat2020synthesizing} proposed a feature synthesis approach for zero-shot object detection and used the mode seeking regularization \cite{mao2019mode} to enhance the diversity of synthesized features. However, these methods do not have sufficient learning capacity for synthesizing region features that are as intra-class diverse and inter-class separable as the real samples, and this is the core problem studied in this work.

\section{Method}
\label{sec:formatting}
\subsection{Problem Definition and Framework Overview}
In ZSD, we have two disjoint sets of classes: seen classes in $\mathcal{Y}^{\rm s}$ and unseen classes in $\mathcal{Y}^{\rm u}$, where $\mathcal{Y}^{\rm s}$ $\cap$ $\mathcal{Y}^{\rm u}$ = $\varnothing$. The training set contains seen objects,
where each image is provided with the corresponding class labels and bounding box coordinates. In contrast, the test set may contain unseen objects only (i.e., the ZSD setting) or both seen and unseen objects (i.e., the GZSD setting). During learning and testing, the semantic word-vectors $\mathcal{W}=\{\mathcal{W}^{\rm s},\mathcal{W}^{ \rm u}\}$ are provided for both seen and unseen classes. The task of ZSD is to learn detectors (parameterized by $\theta$) that can localize and recognize unseen objects corresponding to semantic word-vectors.

Figure~\ref{framwork} shows the proposed overall framework for ZSD. As can be seen,
it contains an object detection module and a domain transformation module. The object detection module is a Faster-RCNN model \cite{ren2016faster} with the ResNet-101 as the backbone \cite{he2016deep}. First of all, we train the Faster-RCNN model with seen images and their corresponding ground-truth annotations. Once the model is obtained, we can use it to extract region features using RPN for seen classes. Second, we train the region feature synthesizer to learn the mapping between semantic word-vectors and the visual features. Then, we use the learned feature synthesizer to generate region features for unseen classes. With these synthesized unseen region features and their corresponding class labels, we can train the unseen classifier for unseen classes. Finally, we update the classifier in the Faster-RCNN model to achieve a new detector for the ZSD task. The overall training procedure is also elaborated in Algorithm \textcolor{red}{1}. 

Notice that the core of our method is how to learn a unified generative model to learn the relationship between visual and semantic domains. Specifically, we design a unified region feature synthesizer for feature synthesizing in real-world detection scenarios, which contains an intra-class semantic diverging component and an inter-class structure preserving component. 


\subsection{The Robust Region Feature Synthesizer}

Given the object feature collection $\mathcal{F}^s$, the corresponding label collection $\mathcal{Y}^s$, and the semantic vector $\mathcal{W}^s$ for seen training data $\mathcal{X}^s$, the goal is to learn a conditional generator ${G}: \mathcal{W}\times \mathcal{Z}\mapsto \mathcal{F}$. That is to say, when we take a class embedding $\textbf{w}\in \mathcal{W}$ and a random noise vector $\textbf{z}\sim\mathcal{N}(0,1)\in \mathbb{R}^d$ sampled from a Gaussian distribution as inputs, we can generate the visual feature $\tilde{\textbf{f}}\in \mathcal{F}$ for the object regions belonging to this class. Then, with the synthesized region features for the unseen classes, we can learn the classifiers for unseen objects. In other words, the generator $G$ learns the mapping between the semantic vectors and the corresponding region features. To learn such a region feature synthesizer, we propose the following learning objective function:
\begin{equation}
\mathop{\min}\limits_{G}\mathop{\max}\limits_{D}\mathcal{L}_{\rm WGAN}+\lambda_{1} \mathcal{L}_{\rm C_s} + \lambda_{2} \mathcal{L}_{\rm S_d} +\lambda_{3} \mathcal{L}_{\rm S_p},
\label{overallLoss}
\end{equation}
where $\mathcal{L}_{WGAN}$ is the Wasserstein GAN loss \cite{arjovsky2017wasserstein} used to enforce the generator to synthesize region features that are aligned well with the distribution of the real region features:
\begin{equation}
\setlength{\abovedisplayskip}{1.8pt}
\begin{split}
\mathcal{L}_{\rm WGAN}=\mathbb{E}[D(\textbf{f}^{\rm s}, \textbf{w}^{\rm s})]-\mathbb{E}[D(\mathbf{\tilde{f}}^{\rm s}, \textbf{w}^{\rm s})]\\
-\lambda \mathbb{E}[(\|\bigtriangledown_{\mathbf{\hat{f}}^{\rm s}}D(\mathbf{\hat{f}}^{\rm s}, \textbf{w}^{\rm s})\|_{2}-1)^{2}],
\end{split}
\label{wganloss}
\end{equation}
where $\mathbf{f}^s$ is the real visual features of the object regions from the seen classes, $\mathbf{\tilde{f}}^{\rm s} = {G}(\textbf{w}^s,\textbf{z})$ denotes the generated visual features conditioned on class semantic vector $\textbf{w}^s\in \mathcal{W}^s$, $\mathbf{\hat{f}}^{\rm s} = \mu \mathbf{\tilde{f}}^{\rm s}$ + (1-$\mu$)$\mathbf{\tilde{f}}^{\rm s}$ with $\mu$ sampled from the uniform distribution $\mu \sim \mathcal{N}(0,1)$, and $\lambda$ is the penalty coefficient.
The discriminator $D$: $\mathcal{F}^{\rm s}$ ($\mathcal{\tilde{F}}^{\rm s}$) $\times$ $\mathcal{W}^{\rm s}$ $\rightarrow$ $[0, 1]$, takes a real region feature $\textbf{f}^{\rm s}$ $\in$ $\mathcal{F}^{\rm s}$ or a synthetic visual feature $\mathbf{\tilde{f}}^{\rm s}$ with the corresponding class-semantic embedding $\textbf{w}^{\rm s}$ as input. It tries to accurately distinguish real visual features from synthetic visual features. In $\mathcal{L}_{\rm WGAN}$,
the first two terms calculate Wasserstein distance, while the third term constrains the gradient of the discriminator $G$ to have unit norm along with the line connecting pairs of real feature $\textbf{f}^{\rm s}$ and synthesized feature $\mathbf{\tilde{f}}^{\rm s}$. $\mathcal{L}_{C_s}$ ensures the generated visual features aligned with the pre-trained classifier on the seen data, which refers to~\cite{hayat2020synthesizing}.

To improve the robustness of the region feature synthesizer, we explore two new learning terms, including $\mathcal{L}_{S_d}$ and $\mathcal{L}_{S_p}$. Specifically, $\mathcal{L}_{S_d}$ is the proposed intra-class semantic diverging loss, which diverges the semantic word-vector of one object category into a set of region visual features. $\mathcal{L}_{S_p}$ is the proposed Inter-class Structure Preserving loss, whose goal is to constrain the separability of the synthesized visual features. $\lambda_1$, $\lambda_2$ and $\lambda_3$ are the weighting hyper-parameters to balance each component. 

\begin{algorithm}[t]
  \renewcommand{\algorithmicrequire}{\textbf{Input:}}
  \renewcommand{\algorithmicensure}{\textbf{Output:}}
  \caption{Training procedure for our framework.}
  \begin{algorithmic}[1]
   \Require
      Training image collection with the corresponding class and bounding box annotations $\{\mathcal{X}^{\rm s},\mathcal{Y}^{\rm s},\mathcal{B}^{\rm s}\}$, Semantic-word vector collection $\mathcal{W}$;
    \Ensure
     Object detector parameters $\theta=\{\theta^{p},\theta^s,\theta^u\}$ ($\theta^{p}$ indicates the parameter for RPN proposal extraction);\\
     $\{\theta^{p},\theta^s\}$ $\leftarrow$ Train object detector on $\{\mathcal{X}^{\rm s},\mathcal{Y}^{\rm s},\mathcal{B}^{\rm s\}}$;\\
     $\mathcal{F}^{\rm s}$ $\leftarrow$ Extract region features from $\mathcal{X}^{\rm s}$ using RPN;\\
     $G$ $\leftarrow$ Train region feature synthesizer on $\mathcal{F}^{\rm s}$ and $\mathcal{Y}^{\rm s}$ by optimizing the loss function in Eq. \ref{overallLoss};\\
     $\mathcal{\tilde{F}}^{\rm u}$ $\leftarrow$ Synthesize region features for unseen classes using the trained $G$ and $\mathcal{W}^{u}$;\\
     $\theta^{\rm u}$ $\leftarrow$ Train unseen object classifier $\theta^{\rm u}$ using $\mathcal{\tilde{F}}^{\rm u}$, $\mathcal{Y}^{\rm u}$;\\
     $\theta$ $\leftarrow$ Update classifier $\theta^{\rm s}$ of the object detection module with $\theta^{\rm u}$;\\
    \Return $\theta$ ;
    \label{implementation-process}
  \end{algorithmic}
\end{algorithm}

\subsection{Intra-class Semantic Diverging}

To enable the model to synthesize diverse visual features, we consider the IntraSD component to diverge the semantic vector of one semantic word-vector into a set of visual features.
Specifically, we conjecture that the above issue could be alleviated by enhancing the influence of noise vectors on the synthetic visual features while preserving the fitness of the synthetic visual features to the class-semantic embeddings. To this end, we propose a novel Intra-class Semantic Diverging loss, where the visual features synthesized from adjacent noise vectors will be pulled closer while those synthesized from distinct noise vectors will be pushed away.

The key design underlying the Intra-class Semantic Diverging loss is how to select the ``positive'' and ``negative'' sample pairs. We design positive samples and negative samples by manipulating the input noise vectors \cite{liu2021divco}. Specifically, given a query noise vector $\textbf{z}$ $\sim$ $\mathcal N(0,1)$, we define a small hyper-sphere with radius $r$ centered at the query noise vector $\textbf{z}$. We random sample a positive query noise vector $\textbf{z}_{+}$ as a vector with the sphere $\textbf{z}_{+}$ $=$ $\textbf{z}$ $+$ $\bm{\rho}$, where $\bm{\rho}$ is a randomly sampled vector from a uniform distribution $\bm{\rho}$ $\sim$ $\mathcal U[-r,r]$. We sample the negative noise vectors as random vectors outside the sphere within a latent space, i.e., $\textbf{z}_{i-} \sim$
$\{\textbf{z}_{i-}|\textbf{z}_{i-} \sim \mathcal N(0,1) \cap |\textbf{z}_{i-} - \textbf{z}| \succ r\}$ for $i=1,\ldots,N$, where $\succ$ is the element-wise greater-than operator. Once these noise vectors are determined, we can define the ``positive'' and ``negative'' samples. For a query visual feature $\mathbf{\tilde{f}}^{\rm s}$ $=$ $G(\mathbf{z},\mathbf{w}^{\rm s})$ synthesized from the noise vector $\mathbf{z}$, we define its ``positive'' sample as $\mathbf{f}^{\rm s}_+$ $=$ $G(\mathbf{z}_+,\mathbf{w}^{\rm s})$ synthesized from the noise vector $\mathbf{z}^{+}$. The $N$ ``negative'' samples synthesized from the sets of noise vectors $\{\mathbf{z}_{i-}\}$ can be defined as $\mathbf{f}^{\rm s}_{i-}$ $=$ $G(\mathbf{z}_{i-},\mathbf{w}^{\rm s})$. The Intra-class Semantic Diverging loss is given by
\begin{equation}
\mathcal{L}_{\rm S_d} = \mathbb{E}[- \log \frac{\mathrm{exp}(\mathbf{\tilde{f}}^{\rm s} \cdot \mathbf{\tilde{f}}^{\rm s}_{\rm +}/\tau)}{\mathrm{exp}(\mathbf{\tilde{f}}^{\rm s} \cdot \mathbf{\tilde{f}}^{\rm s}_{\rm +}/\tau) + \sum_{i=1}^{N}\mathrm{exp}(\mathbf{\tilde{f}}^{\rm s} \cdot \mathbf{\tilde{f}}^{\rm s}_{i-}/\tau)}],
\label{lcr}
\end{equation}
where ``$\cdot$'' is the dot product between two visual feature vectors to measure the cosine similarity and $\tau$ is a temperature scale factor.

\subsection{Inter-class Structure Preserving}
In order to make the synthesized visual features approximate distributions of the real data, meanwhile improving the discrimination of the learned visual features, we further introduce the Inter-class Structure Preserving component into the learning framework. In this learning component, we not only consider the synthesized visual features of different categories, but also pay attention to the real region features extracted by the window proposals, which contain both the positive object proposals, i.e., proposals with the same class as the synthesized feature, and many negative and background proposals.

By doing so, the proposed Inter-class Structure Preserving component has the following merits: 1) The proposed method surpasses the reconstruction error-based loss in the conventional WGAN as it can force the synthesized visual features to be close to other different real visual features of the same category in the window proposal pool. By this way, the synthesized visual features can well approximate the distribution of the real data, facilitating robust one-to-many projection from the semantic word vector to the synthesized region features. 2) By pushing away the visual features from different categories (both in real and synthesized feature space), this learning component can effectively enhance the discrimination of the synthesized visual features.

From the above description, we can observe that the proposed method uses both the synthesized region features and real proposal features to implement the learning process, which essentially constructs a hybrid visual feature pool denoted as $\mathbf{g}=\{{\tilde{\mathbf{f}}^{\rm s},\mathbf{f}^{\rm r},\mathbf{f}^{\rm bg}}\}$, where $\mathbf{f}^{\rm r}$ denotes the real features of the window proposals for different object categories and $\mathbf{f}^{\rm bg}$ indicates the background visual features extracted from the training images. Then, the learning objective function of the proposed Inter-class Structure Preserving component can be written as:
\begin{align}
& \mathcal{L}_{\rm S_p}=\mathbb{E}[- \log \frac{\mathrm{exp}(\mathbf{\tilde{f}}^{\rm s} \cdot \mathbf{\mathbf{g}_{+}}/\tau)}{\mathrm{exp}(\mathbf{\tilde{f}}^{\rm s} \cdot \mathbf{g}_{\rm +}/\tau) + \sum_{j\in\Phi}\mathrm{exp}(\mathbf{\tilde{f}}^{\rm s} \cdot \mathbf{g}_{\rm j}/\tau)}],
\end{align}
where $\Phi=\{{\mathbf g}_{\rm j}\}$ indicates the collection of visual features satisfying $y(\mathbf{g}_j)\neq y(\mathbf{\tilde{f}}^{\rm s})$ in the hybrid visual feature pool, $y(\cdot)$ is the category indicator function, i.e., $y(\tilde{\textbf{f}^s})$ is the class label for visual feature $\tilde{\textbf{f}^s}$. $\mathbf g_{\rm +}$ is the positive examples corresponding to the current synthesized visual feature $\tilde{\textbf{f}^s}$. It can be selected from the synthesized visual features or the object proposals generated by the detector which share the same category label with the current synthesized visual feature $\tilde{\textbf{f}^s}$. Therefore, this Inter-class Structure Preserving loss enables the synthesized visual feature $\tilde{\textbf{f}^s}$ to be close to both the synthesized and real object proposals of the same category, while far apart from all other visual features from different class labels in the hybrid visual feature pool.



\section{Experiment}
\textbf{Datasets:} We evaluate the proposed method on three popular object detection benchmark datasets:  PASCAL VOC 2007+2012 \cite{everingham2010pascal}, MS COCO 2014 \cite{lin2014microsoft}, and DIOR \cite{li2020object}.  The  PASCAL VOC 2007 contains 2501 training images, 2510 validation images, and 5011 test images with 20 categories. PASCAL VOC 2012 contains 5717 training images and 5823 validation images also with 20 categories. MS COCO 2014 contains 82783 training images and 40504 validation images with 80 categories. DIOR contains 5862 training images, 5863 validation images, and 11738 test images with 20 categories. For PASCAL VOC and MS COCO,  we adopt the FastText method \cite{mikolov-etal-2018-advances} to extract the semantic word-vector following \cite{hayat2020synthesizing}. For DIOR, we adopt the Bert model \cite{devlin2018bert} to generate the semantic word-vector.

\textbf{Seen/unseen split:}
We follow the 16/4 seen/unseen split proposed in \cite{Demirel2018ZeroShotOD} on the PASCAL VOC dataset. For MS COCO, we adopt the same setting as \cite{bansal2018zero} to divide the dataset with two different splits: (1) 48/17 seen/unseen split (2) 65/15 seen/unseen split. We divide the DIOR dataset with 16/4 seen/unseen split and the details of the split are provided in the supplementary material. For all the above datasets and splits, we remove all the images of unseen categories from the training set to guarantee that unseen objects will not be available during model training.
\begin{table}[t]
  \centering
  \caption{Comparison of mAP at IoU=0.5, under ZSD and GZSD settings on PASCAL VOC dataset.}
  \vspace{-3mm}
  \renewcommand\tabcolsep{9.0pt}
    \begin{tabular}{lcccc}
    \toprule
    \multirow{2}[4]{*}{Method} & \multirow{2}[4]{*}{ZSD} & \multicolumn{3}{c}{GZSD} \\
\cmidrule{3-5}          &       & S     & U     & HM \\
    \midrule
    SAN \cite{rahman2018zero}  & 59.1  & 48.0    & 37.0    & 41.8 \\
    HRE \cite{Demirel2018ZeroShotOD}  & 54.2  & \textbf{62.4}  & 25.5  & 36.2 \\
    PL \cite{rahman2020improved}   & 62.1  & -     & -     & - \\
    BLC \cite{zheng2020background}  & 55.2  & 58.2  & 22.9  & 32.9 \\
    SU \cite{hayat2020synthesizing}   & 64.9  & -     & -     & - \\
    Ours  & \textbf{65.5}     & 47.1     & \textbf{49.1}     & \textbf{48.1} \\
    \bottomrule
    \end{tabular}%
  \label{tab-voc}%
\end{table}%
\begin{table}[t]
  \centering
  \caption{Class-wise AP and mAP comparison of different methods on unseen classes of PASCAL VOC dataset for ZSD.}
  \renewcommand\tabcolsep{8.0pt}
    \begin{tabular}{lccccc}
    \toprule
    Method & car   & dog   & sofa  & train & mAP \\
    \midrule
    SAN \cite{rahman2018zero}  & 56.2  & 85.3  & \textbf{62.6}  & 26.4  & 57.6 \\
    HRE \cite{Demirel2018ZeroShotOD}  & 55.0    & 82.0    & 55.0    & 26.0    & 54.5 \\
    PL \cite{rahman2020improved}   & 63.7  & 87.2  & 53.2  & 44.1  & 62.1 \\
    BLC \cite{zheng2020background}  & 43.7  & 86    & 60.8  & 30.1  & 55.2 \\
    SU \cite{hayat2020synthesizing}   & 59.6  & 92.7  & 62.3  & 45.2  & 64.9 \\
    Ours  & \textbf{60.1}     & \textbf{93.0}     & 59.7     & \textbf{49.1}     & \textbf{65.5} \\
    \bottomrule
    \end{tabular}%
  \label{clase-wise-AP-voc}%
\end{table}%

\textbf{Evaluation Protocols:} We follow the evaluation strategy proposed in \cite{bansal2018zero, Demirel2018ZeroShotOD}. For PASCAL VOC and DIOR, we utilize mean average precision (mAP) with IoU threshold 0.5 to evaluate the performance. For MS COCO, we utilize mAP with IoU threshold 0.5 and recall@100 with three different IoU thresholds (\emph{i.e.} 0.4, 0.5, and 0.6) as the metric. Furthermore, since the test set consists of seen and unseen images, the performance of GZSD is evaluated by the Harmonic Mean (HM)~\cite{han2021contrastive}.

\textbf{Implementation Details}: Our object detection module adopts the widely-used Faster-RCNN model~\cite{ren2016faster} with the ResNet-101 as the backbone~\cite{he2016deep}. The generator $G$ and discriminator $D$ are both two fully-connected layers with LeakyReLU activation~\cite{maas2013rectifier}. For each unseen class, we synthesize 300/300/500 region features for COCO/DIOR/PASCAL VOC to train the classifiers. The hyperparameter $\lambda_1$ in Eq.~(\ref{overallLoss}) is set to 0.1/0.1/0.01 for COCO/DIOR/PASCAL VOC. Empirically, in the IntraSD component, the trade-off parameter $\lambda_2$ is set to 0.001, the number of negative samples $N$ is set to 10, the temperature coefficient $\tau$ is set to 0.1, and the radius $r$ is set to $10^{-4}$/$10^{-4}$/$10^{-6}$ for COCO/DIOR/PASCAL VOC. For the IntraSP component, the trade-off parameter $\lambda_3$ is set to 0.001 and the temperature coefficient $\tau$ is set to 0.1. 

\begin{table}[t]
  \centering

  \caption{ZSD performance of Recall@100 and mAP with different IoU thresholds on MS COCO dataset.}
  \renewcommand\tabcolsep{3.6pt}
    \begin{tabular}{lccccc}
    \toprule
    \multirow{2}[4]{*}{Method} & \multirow{2}[4]{*}{Split} & \multicolumn{3}{c}{Recall@100} & mAP \\
\cmidrule(r){3-5}   \cmidrule(r){6-6}       &       & IoU=0.4   & IoU=0.5   & IoU=0.6   & IoU=0.5 \\
    \midrule
    SB \cite{bansal2018zero}   & 48/17 & 34.5     & 22.1     & 11.3     & 0.3 \\
    DSES \cite{bansal2018zero}   & 48/17 & 40.2     & 27.2     & 13.6     & 0.5 \\
    TD \cite{li2019zero}   & 48/17 & 45.5     & 34.3     & 18.1     & - \\
    PL \cite{rahman2020improved}   & 48/17 & -     & 43.5     & -     & 10.1 \\
    BLC \cite{zheng2020background}  & 48/17 & 51.3     & 48.8     & 45.0     & 10.6 \\
    Ours  & 48/17 & \textbf{58.1}     & \textbf{53.5}     & \textbf{47.9}     & \textbf{13.4} \\
    \midrule
    PL \cite{rahman2020improved}   & 65/15  & -     & 37.7     & -     & 12.4 \\
    BLC \cite{zheng2020background}   & 65/15 & 57.2     & 54.7     & 51.2     & 14.7 \\
    SU \cite{hayat2020synthesizing}  & 65/15 & 54.4     & 54.0     & 47.0     & 19.0 \\
    Ours  & 65/15 & \textbf{65.3}     & \textbf{62.3}     & \textbf{55.9}     & \textbf{19.8} \\
    \bottomrule
    \end{tabular}%
  \label{zsd-coco}%
\end{table}%
\begin{table}[t]
  \centering

  \caption{Comparison of Recall@100 and mAP at IoU=0.5 over two seen/unseen splits, under GZSD setting on MS COCO dataset.}
  \renewcommand\tabcolsep{4.4pt}
    \begin{tabular}{lccccccc}
    \toprule
    \multirow{2}[4]{*}{Method} & \multirow{2}[4]{*}{Split} & \multicolumn{3}{c}{Recall@100} & \multicolumn{3}{c}{mAP} \\
\cmidrule(r){3-5}     \cmidrule(r){6-8}   &       & S     & U     & HM    & S     & U     & HM \\
    \midrule
    PL \cite{rahman2020improved}   & 48/17 & 38.2     & 26.3     & 31.2     & 35.9     & 4.1     & 7.4 \\
    BLC \cite{zheng2020background}  & 48/17 & 57.6     & 46.4     & 51.4     & 42.1     & 4.5     & 8.2 \\
    Ours  & 48/17 & \textbf{59.7}     & \textbf{58.8}     & \textbf{59.2}     & \textbf{42.3}     & \textbf{13.4}     & \textbf{20.4} \\
    \midrule
    PL  \cite{rahman2020improved}  & 65/15 & 36.4     & 37.2     & 36.8     & 34.1     & 12.4     & 18.2 \\
    BLC \cite{zheng2020background}  & 65/15 & 56.4     & 51.7     & 53.9     & 36.0     & 13.1     & 19.2 \\
    SU \cite{hayat2020synthesizing}  & 65/15 & 57.7     & 53.9     & 55.8     & 36.9     & 19.0     & 25.1 \\
    Ours  & 65/15 & \textbf{58.6}    & \textbf{61.8}     & \textbf{60.2}     & \textbf{37.4}     & \textbf{19.8}     & \textbf{26.0} \\
    \bottomrule
    \end{tabular}%
  \label{gzsd-coco}%
\end{table}%

\subsection{Comparison with the State-of-the-art}
In Table \ref{tab-voc}, we compare with state-of-the-art methods on the PASCAL VOC dataset on ZSD and GZSD settings. We can observe that our method outperforms all the comparison methods in terms of the ZSD setting. Compared with the second-best method SU \cite{hayat2020synthesizing}, our method improves the mAP of ZSD performance from 64.9 $\%$ to 65.5 $\%$. Our method achieves the best performance on unseen classes denoted as ``U'' among all the comparing methods in terms of GZSD setting.
Although our seen performance denoted as ``S" is lower, our method achieves the best performance in terms of ``HM'', which reveals that our method maintains a good balance between seen and unseen classes. This benefits from the robust region feature synthesizer trained with the Intra-class Semantic Diverging component and the Inter-class Structure Preserving component.
We also report the class-wise mAP performance in terms of ZSD setting on the PASCAL VOC dataset in Table \ref{clase-wise-AP-voc}. Our method
achieves the best performance on 3 out of 4 classes, which further demonstrates the superiority of our method on ZSD.
\begin{table*}[t]
  \centering

  \caption{Class-wise AP and mAP comparison of different methods on unseen classes of MS COCO dataset for ZSD.}
      \renewcommand\tabcolsep{5.2pt}
    \begin{tabular}{lcccccccccccccccc}
    \toprule
    65/15 & \multicolumn{1}{c}{airp} & \multicolumn{1}{c}{trai} & \multicolumn{1}{c}{metr} & \multicolumn{1}{c}{cat} & \multicolumn{1}{c}{bear} & \multicolumn{1}{c}{scse} & \multicolumn{1}{c}{frbe} & \multicolumn{1}{c}{snrd} & \multicolumn{1}{c}{fork} & \multicolumn{1}{c}{swic} & \multicolumn{1}{c}{hdog} & \multicolumn{1}{c}{tlet} & \multicolumn{1}{c}{mose} & \multicolumn{1}{c}{tstr} & \multicolumn{1}{c}{hier} & \multicolumn{1}{c}{mAP} \\
    \midrule
    PL \cite{rahman2020improved}   & \multicolumn{1}{c}{20} & \multicolumn{1}{c}{48.2} & \multicolumn{1}{c}{0.6} & \multicolumn{1}{c}{28.3} & \multicolumn{1}{c}{13.8} & \multicolumn{1}{c}{\textbf{12.4}} & \multicolumn{1}{c}{\textbf{21.8}} & \multicolumn{1}{c}{15.1} & \multicolumn{1}{c}{8.9} & \multicolumn{1}{c}{8.5} & \multicolumn{1}{c}{\textbf{0.9}} & \multicolumn{1}{c}{5.7} & \multicolumn{1}{c}{0.0} & \multicolumn{1}{c}{\textbf{1.7}} & \multicolumn{1}{c}{0.0} & \multicolumn{1}{c}{12.4} \\
    SU  \cite{hayat2020synthesizing}  & \multicolumn{1}{c}{10.1} & \multicolumn{1}{c}{48.7} & \multicolumn{1}{c}{1.2} & \multicolumn{1}{c}{64.0} & \multicolumn{1}{c}{\textbf{64.1}} & \multicolumn{1}{c}{12.2} & \multicolumn{1}{c}{0.7} & \multicolumn{1}{c}{28} & \multicolumn{1}{c}{16.4} & \multicolumn{1}{c}{19.4} & \multicolumn{1}{c}{0.1} & \multicolumn{1}{c}{\textbf{18.7}} & \multicolumn{1}{c}{1.2} & \multicolumn{1}{c}{0.5} & \multicolumn{1}{c}{0.2} & \multicolumn{1}{c}{19.0} \\
    Ours  &\textbf{20.8}      & \textbf{53.0}     & \textbf{1.3}      & \textbf{64.3}      &55.5       & 11.6      & 0.4      & \textbf{31.3}      & \textbf{18.0}      &\textbf{20.3}       &0.1       & 15.2      &\textbf{4.2}       &0.5       &\textbf{0.6}       &\textbf{19.8}  \\
    \bottomrule
    \end{tabular}%
  \label{clase-wise-AP-coco}%
\end{table*}%

In Table \ref{zsd-coco}, we compare our method with the state-of-the-art methods on MS COCO dataset over two splits. For the 47/17 split, our method outperforms all the compared methods by a large margin in terms of both Recall@100 and mAP measurements. Compared with the second-best method BLC \cite{zheng2020background}, our method improves the Recall@100 by 9.6 $\%$ and mAP by 26.4 $\%$ at IoU=0.5. For the 65/15 split, we can observe that our method also achieves a significant performance gain, which improves the Recall@100 and mAP of method SU \cite{hayat2020synthesizing} from 54.0 $\%$ and 19.0 $\%$ to 62.3 $\%$ and 19.8 $\%$ at IoU=0.5.

In Table \ref{gzsd-coco}, we compare our method with other methods under the GZSD scenario, which is more realistic and challenging. Our method outperforms all the comparison methods over two splits in terms of all the metrics.  Compared with the second-best method BLC \cite{zheng2020background}, our method improves the ``HM" performance in Recall@100 and mAP from 51.4 $\%$ and 8.2 $\%$ to 59.2 $\%$ and 20.4 $\%$ under the split 48/17. For the 65/15 split, our method improves the ``HM" performance achieved by the second-best method SU \cite{hayat2020synthesizing} from 55.8 $\%$ and 25.1 $\%$ to 60.2 $\%$ and 26.0 $\%$. This ``HM" performance gain proves that our region feature synthesizer can synthesize robust features for unseen classes.

We report clas-wise AP of our method in Table \ref{clase-wise-AP-coco} for 65/15 split. Our method achieves the best performance on 9 out of 15 classes and comparable performance on other classes. Since other methods did not report their class-wise AP results on the 48/17 split, we show our class-wise AP in the supplementary material, individually.

\begin{table}[t]
  \centering

  \caption{Comparison of mAP at IoU=0.5, under ZSD and GZSD settings on DIOR dataset.}
  \renewcommand\tabcolsep{9.0pt}
    \begin{tabular}{lcccc}
    \toprule
    \multirow{2}[4]{*}{Method} & \multirow{2}[4]{*}{ZSD} & \multicolumn{3}{c}{GZSD} \\
\cmidrule{3-5}          &       & S     & U     & HM \\
    \midrule
    PL \cite{rahman2020improved}   & 0.4   & 4.3   & 0.0     & 0.0 \\
    BLC \cite{zheng2020background}  & 1.1   & 6.1   & 0.4   & 0.8 \\
    SU \cite{hayat2020synthesizing}   & 10.5  & \textbf{30.9}  & 2.9   & 5.3 \\
    Ours  & \textbf{11.3}  & \textbf{30.9}  & \textbf{3.4}   & \textbf{6.1} \\
    \bottomrule
    \end{tabular}%
  \label{tab-dior}%
\end{table}%
To further verify the effectiveness of our method, we conduct the experiment on the DIOR dataset, which is the first attempt for implementing zero-shot object detection in remote sensing imagery. We re-implement the
state-of-the-art zero-shot object detection methods based on their released codes on the DIOR dataset for comparison in Table \ref{tab-dior}. Compared with the second-best method SU \cite{hayat2020synthesizing}, the ``ZSD", ``U", and ``HM" are improved from 10.5 $\%$, 2.9 $\%$, and 5.3 $\%$ to 11.3 $\%$, 3.4 $\%$, and 6.1 $\%$, respectively. We also achieve the same S performance as method SU \cite{hayat2020synthesizing}. Due to space limitation, the concrete class-wise AP scores will be reported in the supplementary material.

\subsection{Ablation Study}
\label{ablation_study}
To provide further insight into our method, we conduct ablation studies on the PASCAL VOC dataset to analyze the contributions of each component in our method. In Table \ref{ablation}, We report the ZSD and GZSD performance in terms of mAP metric at IoU 0.5. $\mathcal{L}_{\rm b}$ contains the $\mathcal{L}_{\rm WGAN}$ and $\mathcal{L}_{\rm C_s}$, which can be regarded as our baseline method. $\mathcal{L}_{\rm S_{ps}}$ means the hybrid visual feature pool $g$ only contains the synthesized visual features $\tilde{\mathbf{f}}_{\rm s}$. ``$\checkmark$" denotes the model with the corresponding component.

\begin{table}[t]
  \centering

  \caption{Performance of ablation studies under the ZSD and GZSD settings, measured by the mAP on PASCAL VOC dataset.}
  \renewcommand\tabcolsep{5.7pt}
    \begin{tabular}{cccccccc}
    \toprule
    \multirow{2}[4]{*}{$\mathcal{L}_{\rm b}$} & \multirow{2}[4]{*}{$\mathcal{L}_{\rm S_d}$} & \multirow{2}[4]{*}{$\mathcal{L}_{\rm S_{ps}}$} & \multirow{2}[4]{*}{$\mathcal{L}_{\rm S_p}$} & \multirow{2}[4]{*}{ZSD} & \multicolumn{3}{c}{GZSD} \\
\cmidrule{6-8}          &       &       &       &       & S     & U     & HM \\
    \midrule
    $\checkmark$     &       &       &       & 62.1     &47.1     &45.9      & 46.5  \\
    $\checkmark$     & $\checkmark$     &       &       &64.0       &47.1       &48.3       & 47.7 \\
    $\checkmark$     & $\checkmark$     & $\checkmark$     &       &64.7       &47.1       &48.7       &47.9  \\
    $\checkmark$     & $\checkmark$     &      & $\checkmark$  &\textbf{65.5}     & \textbf{47.1}      & \textbf{49.1}      & \textbf{48.1} \\
    \bottomrule
    \end{tabular}%
  \label{ablation}%
\end{table}%
\begin{figure}
  \centering
  \begin{subfigure}{0.48\linewidth}
    \includegraphics[width=1\linewidth]{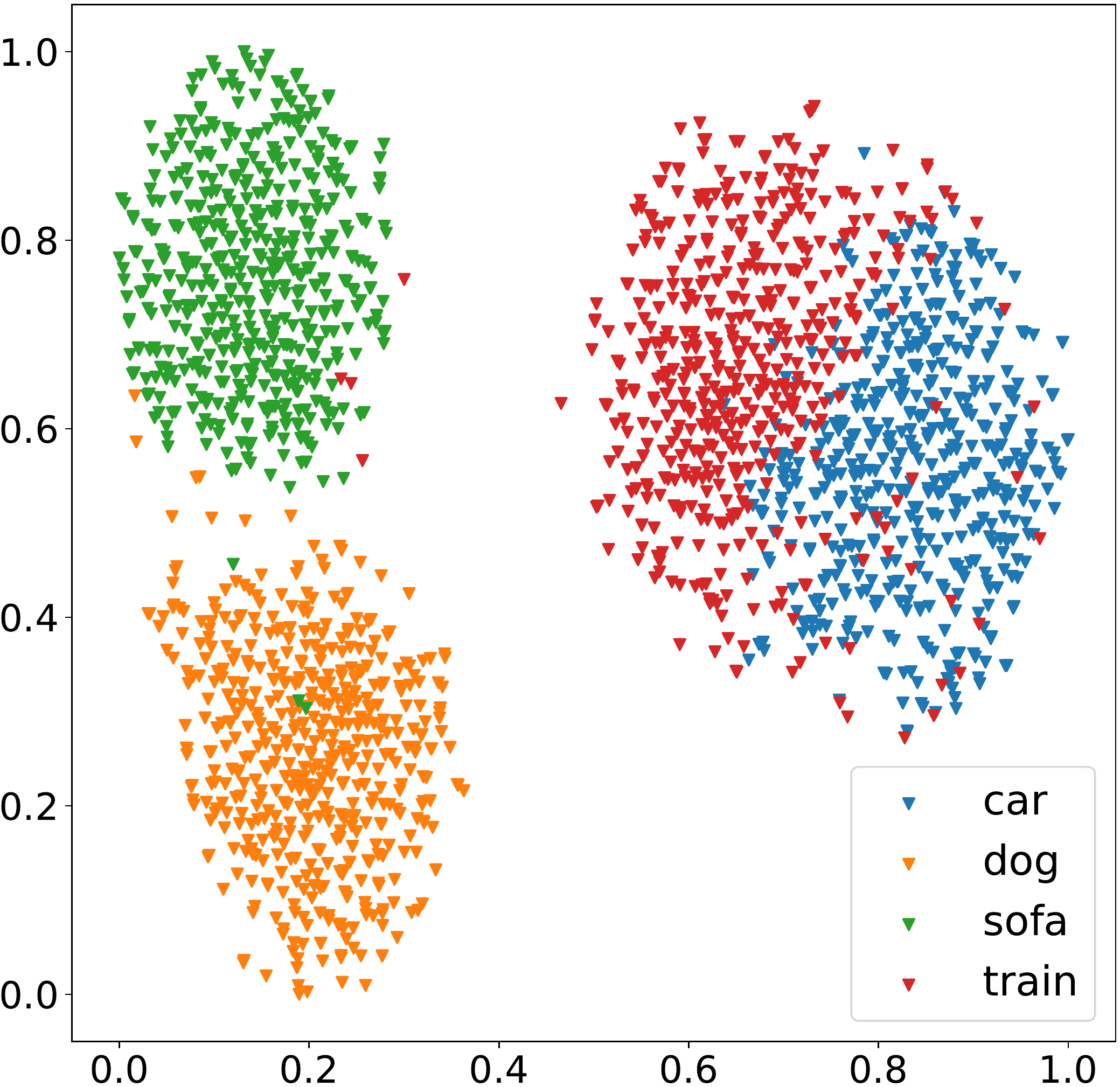}
    \caption{Baseline} 
    \label{seen}
  \end{subfigure}
  \hfill
  \begin{subfigure}{0.48\linewidth}
    \includegraphics[width=1\linewidth]{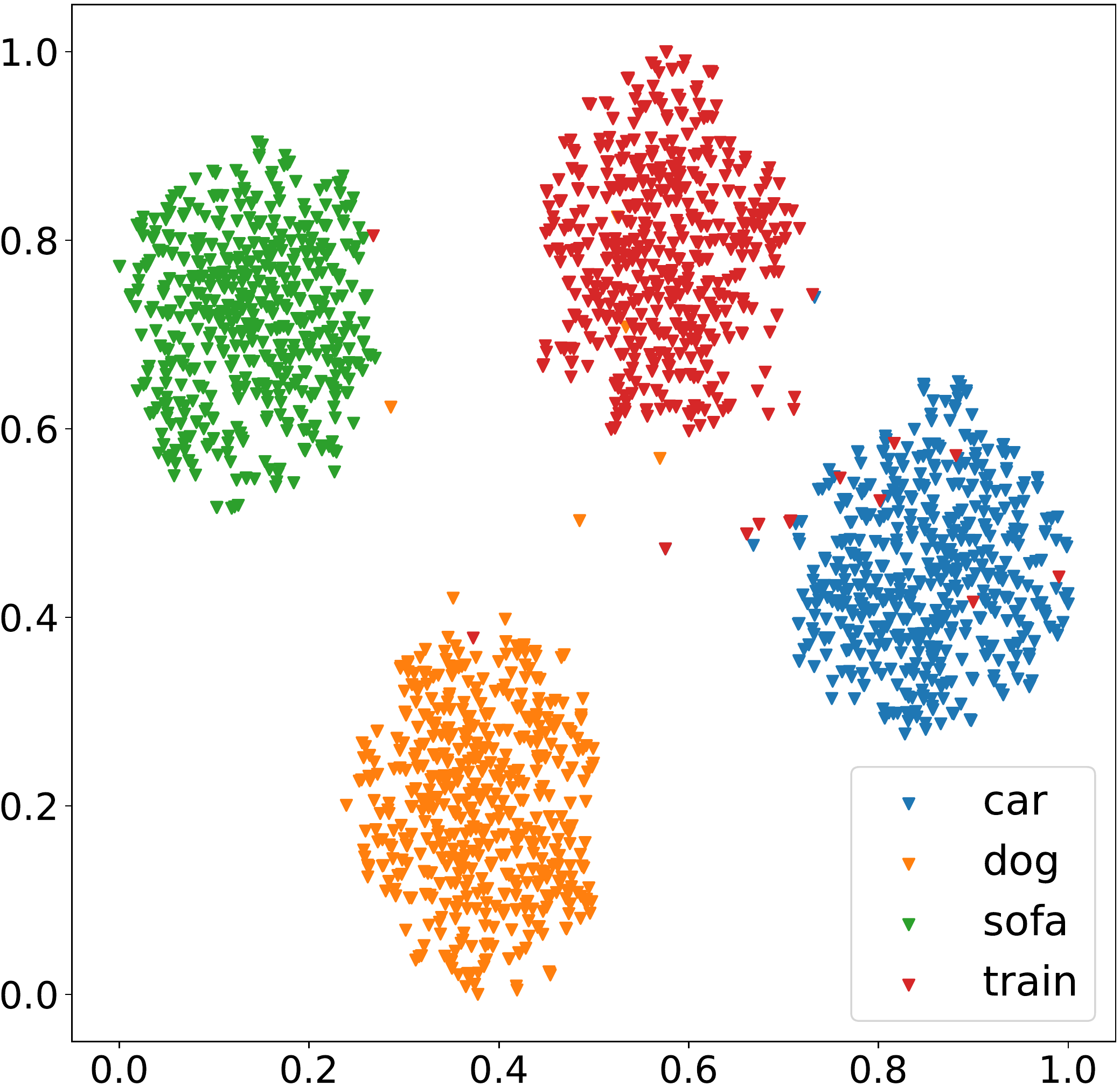}
    \caption{Ours}
    \label{unseen}
  \end{subfigure}
  \caption{The t-SNE \cite{van2008visualizing} visualization of the synthesized region features for the unseen classes on PASCAL VOC dataset.}
  \label{fig:short}
\label{tsne}
\vspace{-3.0mm}
\end{figure}

\textbf{IntraSD Component Analysis.} We first analyze the effectiveness of the proposed IntraSD component. To verify its contributions, we compare the performances of our baseline model and the variant by adding the IntraSD during training. We can observe that the ``ZSD" performance and the ``U" performance of GZSD all have been improved significantly from 62.1 $\%$ and 45.9 $\%$ to 64.0 $\%$ and 48.3 $\%$, respectively. The HM performance is improved from 46.5 $\%$ to 47.7 $\%$. These large performance gains demonstrate the effectiveness of the proposed IntraSD component in the ZSD model, which can encourage our generator to synthesize more diverse visual features for unseen classes. The seen performances of GZSD do not obtain performance gains since the parameters of the classifier for seen class is fixed.
\begin{figure*}[t]
  \centering
   \includegraphics[width=1.0\linewidth]{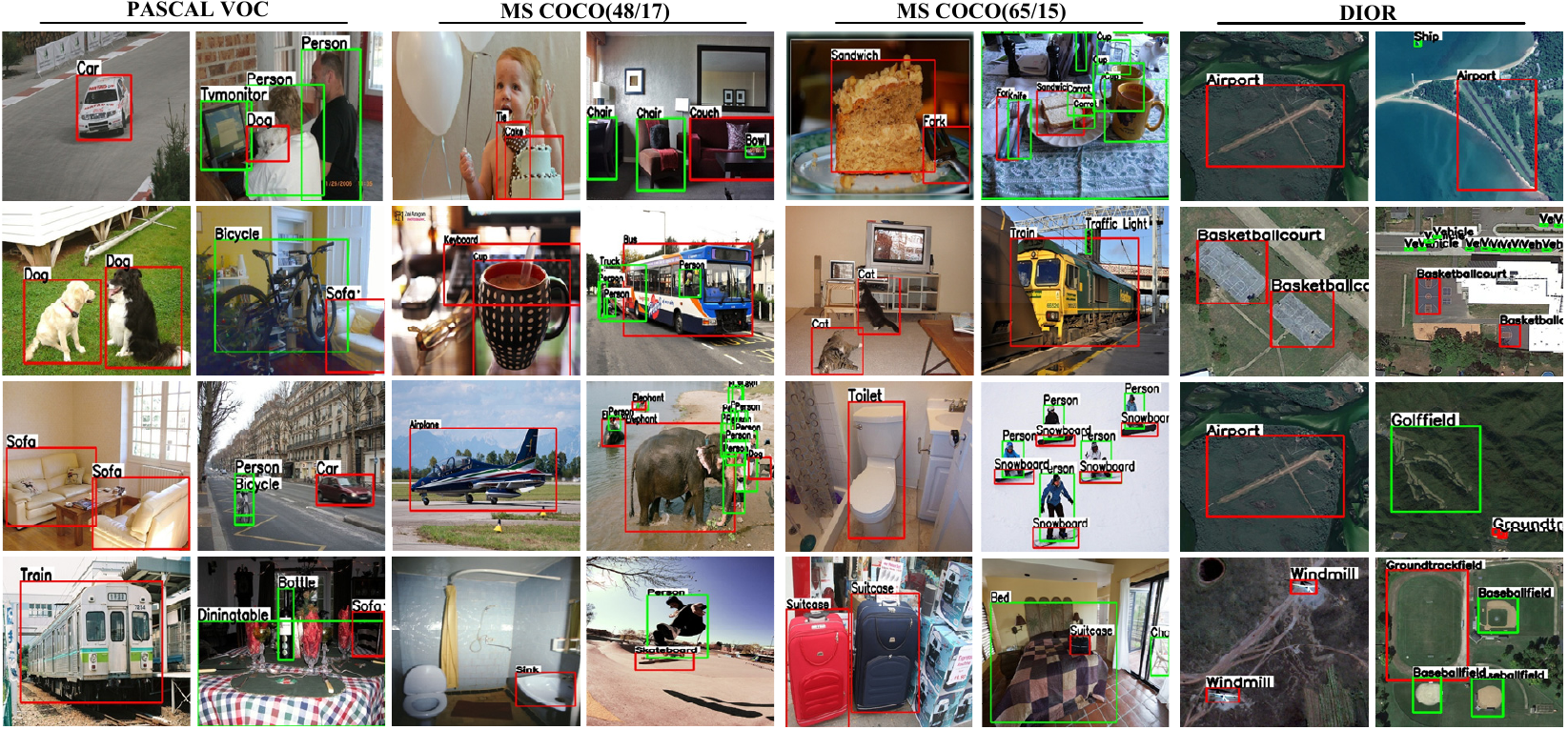}
   \caption{Qualitative results on PASCAL VOC, MS COCO (48/17 and 65/15) and DIOR datasets. For each dataset, the first column and second column are the results of ZSD and GZSD, respectively. Seen classes are shown with green and unseen with red.}
   \label{detection-results}
   \vspace{-4.2mm}
\end{figure*}

\textbf{InterSP Component Analysis}.  To verify the effectiveness of the InterSP component, we first add the $\mathcal{L}_{\rm S_{ps}}$ component on top of the $\mathcal{L}_{\rm b}$ and $\mathcal{L}_{\rm S_d}$. As a result, the mAP measurements of ``ZSD", ``U", and ``HM" are improved from 64.0 $\%$, 48.3 $\%$, and 47.7 $\%$ to 64.7 $\%$, 48.7 $\%$, and 47.9 $\%$. Secondly, we further add the $\mathcal{L}_{\rm S_p}$ component on top of the $\mathcal{L}_{\rm b}$ and $\mathcal{L}_{\rm S_d}$. The mAP measurement of ``ZSD", ``U", and ``HM" are improved from 64.0 $\%$, 48.3 $\%$, and 47.7 $\%$ to 65.5 $\%$, 49.1 $\%$, and 48.1 $\%$. These two comparisons demonstrate that our InterSP component can improve the discrimination of the learned visual features. Compared with the method variant $\mathcal{L}_{\rm S_{ps}}$, the variant with $\mathcal{L}_{\rm S_p}$ gains an absolute improvement of 1.2 $\%$, 0.8 $\%$, and 0.4 $\%$. This phenomenon indicates that the real features of the window proposals play an important role in our InterSP module, since it contains the ground-truth positive object proposals and many background negative proposals.

\subsection{Qualitative Results}
\textbf{Feature visualization} In Fig \ref{tsne}, we conduct the t-SNE \cite{van2008visualizing} visualization for the synthesized region features of the unseen classes on the PASCAL VOC dataset. The visual feature distributions corresponding to our baseline model and the proposed model have been illustrated in Fig \ref{tsne}(a) and Fig \ref{tsne}(b). Features from similar classes (car and train in Fig \ref{tsne}(a)) are confused with each other due to high similarity in their semantic space, which may lead to miss-classify these similar classes. The synthesized features in Fig \ref{tsne}(b) have obvious segregated clusters.
This verifies that our synthesizer is robust in synthesizing intra-class diverse and inter-class discriminative region features, which benefits learning a more discriminative classifier to improve the detection performance for ZSD.

\textbf{Detection Results}. To further show the effectiveness of our method, we show the detection results of our method on PASCAL VOC, MS COCO, and DIOR datasets in Fig \ref{detection-results}. For the ZSD setting, the images only contain unseen objects. For the GZSD setting, the images may contain seen and unseen objects together. The qualitative results prove the effectiveness of our method in detecting seen and unseen objects simultaneously in the challenging scenarios.


\section{Conclusion and Limitation}
In this work, we focus on ZSD task by addressing the challenge of synthesizing robust region features for unseen objects.
Specifically, we propose a novel ZSD framework by constructing a robust region feature synthesizer, which includes the IntraSD and InterSP components. The IntraSD realizes the one-to-more mapping to obtain diverse visual features from each class semantic vector, preventing miss-classifying the real unseen objects as image backgrounds. The InterSP component improves the discrimination of the synthesized visual features by make full use of both synthesized and real region features from different object categories. Extensive experimental results demonstrate  that our method is superior to state-of-the-art approaches for ZSD.

\vspace{1mm}
\noindent \textbf{Limitation.} One major limitation in this study is that the proposed method is based on the two-stage object detector, e.g., Faster-RCNN \cite{ren2016faster}, whose detection speed is relatively slow. We hope to integrate our method into some one-stage object detector, e.g., YOLOv5~\cite{jocher2021ultralytics}, to further improve the detection speed in the future.


{\small
\bibliographystyle{ieee_fullname}
\bibliography{egbib}
}

\end{document}